# Automated Detection of hidden Damages and Impurities in Aluminum Die Casting Materials and Fibre-Metal Laminates using Low-quality X-ray Radiography, Synthetic X-ray Data Augmentation by Simulation, and Machine Learning

## Stefan Bosse[1,2], Dirk Lehmhus[3]

[1]University of Bremen, Dept. Mathematics & Computer Science, Bremen, Germany
[2]University of Siegen, Dept. Mechanical Engineering, Siegen, Germany
[3]Fraunhofer IFAM, Bremen, Germany

**Abstract. Detection and characterization of hidden defects, impurities, and damages in layered composites like Fibre laminates, e.g., Fibre Metal Laminates (FML), as well as in monolithic materials, e.g., aluminum die casting materials, is still a challenge. This work discusses methods and challenges in data-driven modeling of automated damage and defect detectors using X-ray single- and multi-projection (CT) images. Three main issues are identified: Data and feature variance, data feature labeling (for supervised machine learning), and the missing ground truth. It will be shown that only simulation of data can deliver a ground truth data set and accurate labeling. Noise has significant impact on the feature detection and will be discussed. Data-driven feature detectors are implemented with semantic pixel- or z-profile Convolutional Neural Networks and LSTM Auto-encoders. Data is measured with three different devices: A low-quality and low-cost (Low-Q), a mid- and a high-quality (microCT, Mid-/High-Q) device. The goals of this work are the training of robust and generalized feature detectors with synthetic data and the transition from High- and Mid-Q laboratory measuring technologies towards in-field usable technologies and methods.**

## 1. Introduction

Detection and characterization of hidden defects, impurities, and damages in layered composites like Fibre laminates, e.g., Fibre Metal Laminates (FML), as well as in monolithic materials, e.g., aluminum die casting materials, is still a challenge. Commonly, Guided Ultrasonic Waves (GUW) or X-ray imaging are used to detect hidden damages. X-ray imaging can be divided into two-dimensional single.projection transmission or reflection and multi-projection three-dimensional tomography imaging using reconstruction algorithms to compute a three-dimensional view from slice images.

Damages or defects can be classified roughly in layer delaminations, extended cracks, micro cracks (fibres and solid material layer), deformations, and impurities during manufacturing. Detection of such kind of damages and defects by visual inspection is a challenge, even using 3D CT data, and moreover using single 2D transmission images. For damage characterization, micro-focus CT X-ray scanner are used, providing a high resolution below 100 µm, but with the disadvantage of high scanning times (up to several hours) [5].

Anomaly detectors based on advanced data-driven Machine Learning methods can be used to mark Regions-of-Interest (ROI) in images automatically (feature selection process). ROI feature extraction is the first stage of an automated damage diagnostic system providing damage detection, classification, and localization. But data-driven methods require typically a sufficient large set of training examples (with respect to diversity and generality), which cannot be provided commonly in engineering and damage diagnostics (e.g., an impact damage can only be "created" one time and is not reversible).

In this work, the challenges, limits, and detection accuracy of automated ROI damage feature detection from low-quality and low-resolution 2D X-ray image data using data-driven anomaly detectors are investigated and evaluated. In addition to experimental data, X-ray simulation is used to create an augmented training and test data set. Simulated and experimental X-ray data are compared. The simulation is carried out with the gvirtualxray [2,3] software  It is based on the Beer-Lambert law to compute the absorption of light (i.e. photons) by 3D objects (here polygon meshes). Additionally, X-ray ray-tracing by the the x-ray projection simulator [4] software is used for comparison.

Suitable data-driven anomaly detectors estimating and marking the ROI candidates of damage areas are Convolutional Neural Networks trained supervised (i.e., using manually feature labeled data), either used as a pixel-based feature classifier (Point-Net) or as a region-based proposal network (Region-based CNN, R-CNN, Fast R-CNN, Region-proposal networks) [1].

The generated knowledge and the image data collected would further accelerate the development in the field of autonomous SHM of the composite structures which would further reduce the safety risks and total time associated with structural integrity assessment.

Feature detection and marking in measuring images can occur on different levels:

- Region-of-Interest Search

- Feature Maps

- Damage and defect classification

- Damage and defect localization

- Global statistical aggregates (e.g., pore density, distribution)

Either classical numerical and model-based algorithms (e.g., edge detection using a Soebel filter or Canny detectors) or data-driven models are used for feature marking („Machine Learning").

The primary goal of this work is automated damage, defect, and impurity detection in materials including composites using single X-ray projection images (from Low-Q or Mid-Q devices) and data-driven feature marking models (e.g., deploying Convolutional Neural Networks).

Spatially resolved inspection and testing of structures requires image-based measuring methods. Non-destructive testing (NDT) of metal-based structures can exploit different imaging methods, mainly:

- X-ray Radiography (single projection) and Computer Tomography (CT, multi-projection)

- Guided Ultrasonic Waves (GUW) and Ultrasonic Sonography

Homogeneous as well as Composite Materials can be tested, but reflection and diffraction can have a significant impact on image quality (especially in the case of guided US waves!

Detection of hidden damages, defects, and impurities (e.g., pores) is still a challenge. X-ray images pose typically low contrast if the density of defects si close to the host material.

Different specimens, structure geometries, materials, and defects are considered in this work! They pose different coincidence between material and image features:

- Homogeneous aluminum die casting plates (150x30 mm) with gas pore defects

- Composite Fibre Metal Laminate plates (FML, aluminum and PREG layers, 50 x 50 mm) with impact damages posing layer delaminations, deformation, cracks, and kissing bond defects.

Data-driven models require data and the data must contain a sufficient statistical variance and distribution of features to be detected. That's the first issue with most engineering data. Additionally, supervised data modeling requires accurately labeled (annotated) strong feature examples, commonly not available, and being the second issue and downfall in data-driven modeling. This can be summarized by the missing ground truth issue.

The overall road-map of our work is as follows:

1. Analyze real samples for characteristics of pore size distributions, addressed in this paper.

2. Generation of different forms of pore size distributions with identical features in each case directly in an FEM model, or in a CAD model for the purpose of conversion into an FEM model. The FEM model should represent a limited volume, a REV (representative volume element).

3. Virtual testing of the different REVs, determination of (a) properties and (b) distributions of the same for pore size distributions with certain more general, detectable properties such as pore size distribution.

4. Transfer of the findings to (macroscopic) FEM simulations of castings - i.e. no representation of the pores in the FEM model at this level (as is the case with the REV), but mapping of the (now approximately known) variability of the local properties depending on knowledge of local pore size distributions, e.g. from simulation (in any case, this would probably require a stochastic simulation approach).

The next sections introduce our concept of the deployment of a data-driven feature marking model applied to real measured X-ray images, but entirely trained (created) using synthetic image data computed from CAD models, overcoming the missing griund truth issue.

## 2. Concept

The principle concept combines data from a real X-ray radiography (or CT) device and an X-ray image simulator, as shown in Fig. 1. There are different X-ray measuring devices are used, differing in resolution, Field of View (FOV), and achievable signal-to-noise ratios, discussed later. We trained a data-driven model, here a CNN semantic pixel classifier, by using synthetic data from simulation only, finally applying this feature detector model to real measured data.

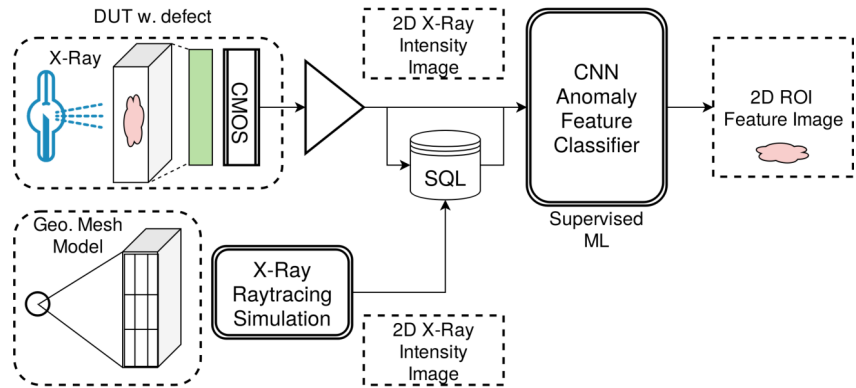

*Fig. 1. Principle concept fusing real and simulated X-ray image data to train a feature or anomaly classifier model.*

| Feature/Device Class | High-Q | Mid-Q | Low-Q |
|---|---|---|---|
| Single Projection | yes | yes | yes |
| Mult-Projection (Rotation/CT) | yes | yes | maybe |
| X-ray Tube Focal Diameter | 5µm | 0.8mm | 0.8mm |
| X-ray Voltage/Current | -120 kV/1 mA | -120 kV/10 mA | -60 kV/1 mA |
| Detector | 2000x2000 20 µm Direct Sci./Imag. Microscope | 1000x1000 200 µm Direct Sci. | 2000x1000 3/40 µm Screen/Imaging |
| Digital Resolution [Bits] | 16 | 16 | 12 |
| Sampling Time | 500 ms-10 s | 100 ms-1 s | 5-10 s |
| Distance Object/Source | 5-10 cm | 20-70 cm | 20-30 cm |
| Signal-Noise Ratio (SNR) | High | Mid | Low |
| Costs | 1000 k€ (Zeiss) | 500 k€ (IFAM) | 1 k€ (Bosse) |

*Tab. 1. Different measuring device classes used in this work*

The advanced concept introduced in this work incorporates multiple measuring devices and experiments, as shown in Fig. 2. The used measuring devices are summarized in the Tab. 1, abbreviated by Low-Q, Mid-Q, and High-Q.

The High-Q Micro-CT device is used only as a reference and to get a basic set of defect parameters and statistical characteristics from real specimens. High resolution reconstructed CT volumes are analyzed and defect (pore) parameters are extracted by hand using an image measuring tool (ImageJ). All data is stored in a SQL data-base.

Overall we perform a data and method fusion, as illustrated in Fig. 3, to derive robust feature marking models applied to single projection X-ray images. the method space covers different data processing algorithms and measuring technologies. The data-driven feature marking model creates latent features that are used for the final defect characterization.

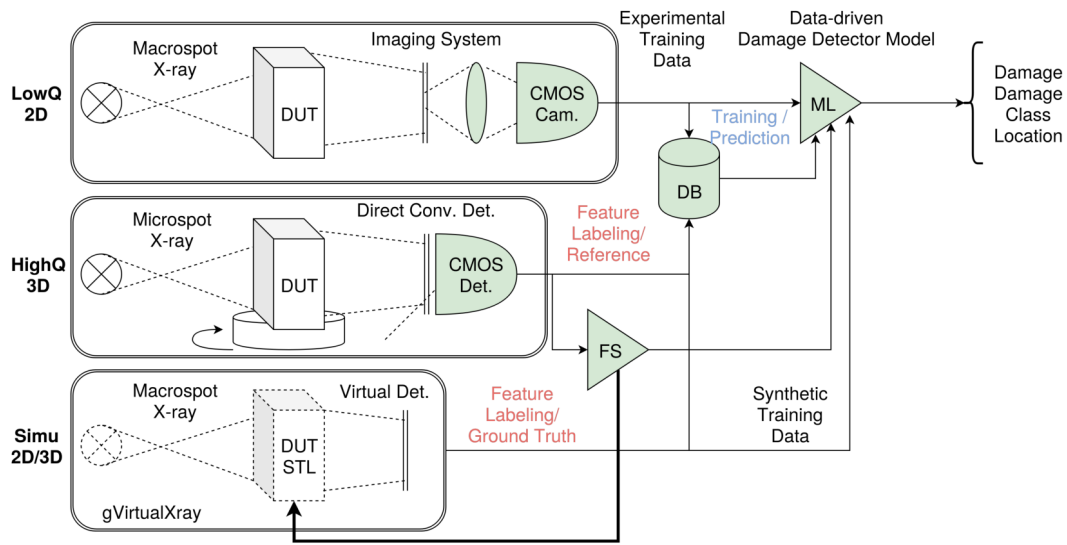

*Fig. 2. Advanced concept (FS: Feature Selection, DB: Data-base, ML: Data-driven Machine Learning, DUT.*

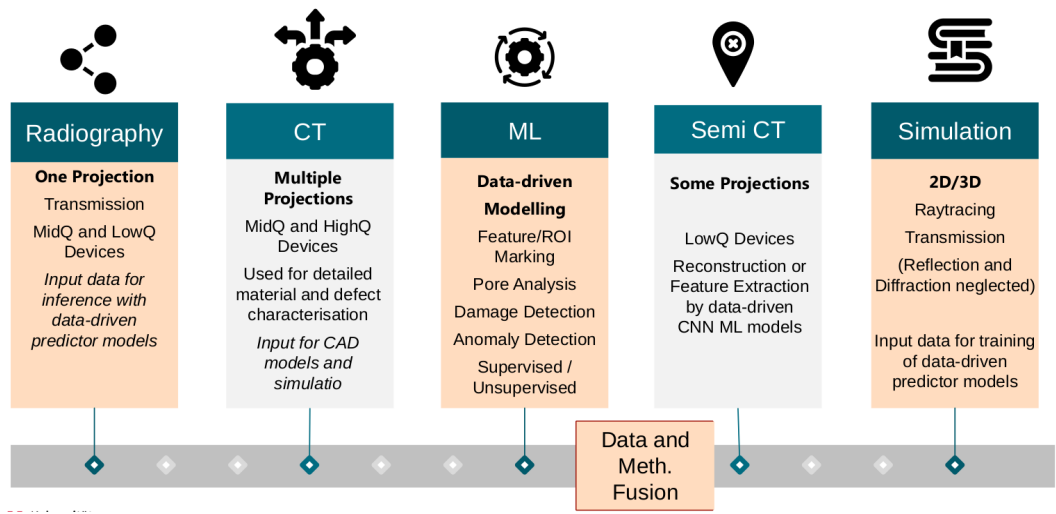

*Fig. 3. Data and method fusion for advanced feature detection in X-ray images*

## 3. Taxonomy and Capabilities of NDT Methods

Fig. 4 illustrates the comparison of two prominent measuring methods suitable for defect and material inhomogeneity detection and characterization. They differ in the measuring wave interaction with materials and defects. Ultrasonic wave interaction is mainly related to material density variations, dispersion, reflection, and mode conversions. So the sensor bases commonly on reflection, providing time-of-light (1D/2D) or sonogram features (3D). X-ray wave interaction bases mainly on material density variation. Reflection and scattering is a minor effect, but must be considered in metals. The measuring method bases mainly on transmission and absorption with respect to the material density along a X-ray path. X-ray radiography and topography provide always 2D sensor data, but tomography captures multiple projection images (commonly by rotating the specimen). 3D tomography image stacks are computed by reconstruction and filtering algorithms from rotated projections.

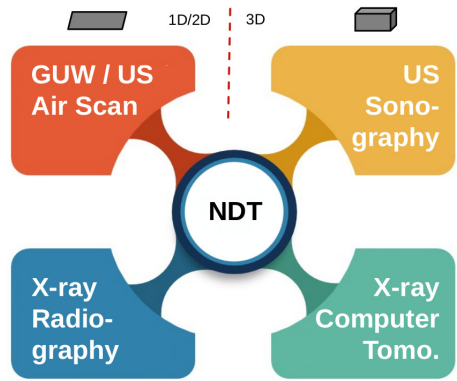

*Fig. 4. Comparison chart of two prominent NDT measuring methods: X-ray and Ultrasonic.*

X-ray images can be simulated with high accuracy with respect to real measured images.X-ray images enable direct interpretation and feature detection (e.g., damages), but, not all features are directly visible (low contrast) and need to be intensified (improving contrast/SNR by algorithms). In contrast, Ultrasonic signals cannot be simulated with high accuracy with respect to real measured images; there is a large reality gap!

Typical feature classes that can be detected by different methods with different sensitivity and accuracy are: Damages of regions of materials or layers of a composite materials, defects (including manufacturing defects), inhomogeneities (material density, geometric characteristics), pores, delaminations, cracks. Delaminations cannot be detected by X-ray imaging directly (no density change along a X-ray path), but can be identified by US methods (except delamniations as a secondary effects from deforma-tion).

## 4. Specimens and Experiments

We consider two use-cases to demonstrate the capability and issues of data-driven modelling using synthetic simulated data only:

1. Pore detection (feature marking) from single frontal LowQ X-ray projections using a Convolutional Neural Network

2. Damage or anomaly feature marking in 3D CT reconstructed HighQ image volumes using a Convolutional Neural Network

Therefore, we used to different specimen classes, with examples shown in Fig. 5:

1. Aluminum die casting plates with pores, Fraunhofer IFAM Bremen (Dirk Lehmhus)

2. GLARE Fibre Metal Laminate plates (5 layers) with impact damages , DFG research group 3022 (Bremen, Hamburg, Braunschweig, Siegen)

Different measurements were done, respectively:

1. Low-Q X-ray (Bosse),single and multi-projection images, 55 kV, 1mA, 1920 x 1080 pixels, Imaging detector with CMOS sensor (Sony IMX290), effective 40 µm resolution, a detailed description can be found in [11];

2. Mid-Q X-ray (IFAM), single and multi-projection images (CT, 400/800 projections) 60 kV, 2 mA, 1000 x 1000 pixels SSD, 200µm resolution;

3. High-Q X-ray (Zeiss Xradia µCT) multi-projection images (CT, 800 projections), 110 kV, 1 mA, 2000 x 2000 pix. SSD, 20 µm resolution.

The three measuring device classes are summarized in Tab. 1.

The lack of data variance is the first major challenge to be solved by creating synthetic data retrieved from simulation. But things differ with respect of the two applications. Regarding the pore analysis, each pore is an individual sample instance, and a sufficient number of plate specimens are available, finally giving us more than 10000 individual data instances. In contrast to the FML damage detection application. Here we have only a few plates, each providing only one primary impact damage with some secondary damages (mainly delaminations and cracks). The different situations are summarized in Fig. 6. In this work a semantic pixel classifier is used for feature marking. From the model point of view, each pixel (and neighboring pixels) of an X-ray image is a sample instance.

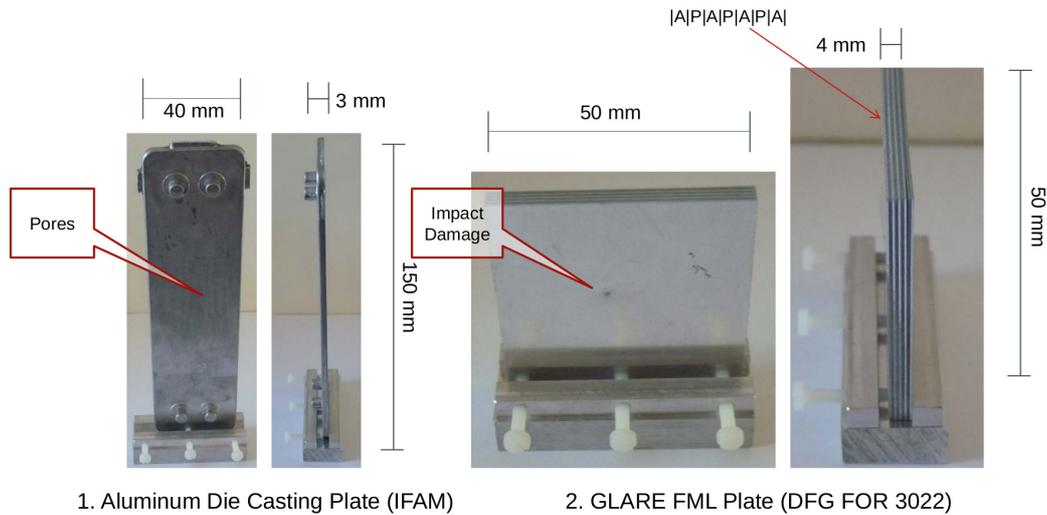

*Fig. 5. Two classes of specimens used in this work: 1. Die casted aluminum plates 2. Fibre-Metal Laminate plates*

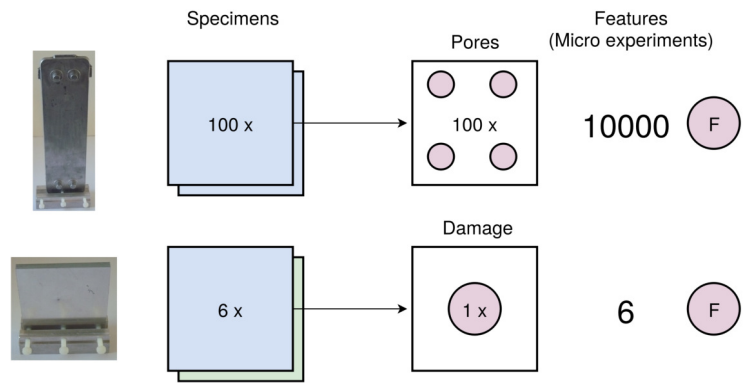

*Fig. 6. The first challenge: The lack of data variance and scale of sample instances*

The measured data was analyzed by manual inspection using image processing software, creating training data for data-driven feature marking models, finally applied to the measured or reconstructed image data providing automated feature marking of defects and damages. The methods are introduced and discussed in the following Section.

## 5. Taxonomy of Feature Detection

This work covers different defect and damage classes, materials, and measuring technologies, summarized in Fig. 7. We are sharing basically the same ML methods and algorithms for both use cases. In the case of X-ray intensity images, a semantic pixel classifier is applied to the input image creating a feature map, and in the case of X-ray CT density image stacks, a signal structure classifier is applied to z-profile slices creating a x-y feature map (seen from the top of the specimen surface).

Currently, the output of the ML models are intermediate feature maps that have to be post-processed, e.g., using feature map pixel clustering and geometric shape fitting.

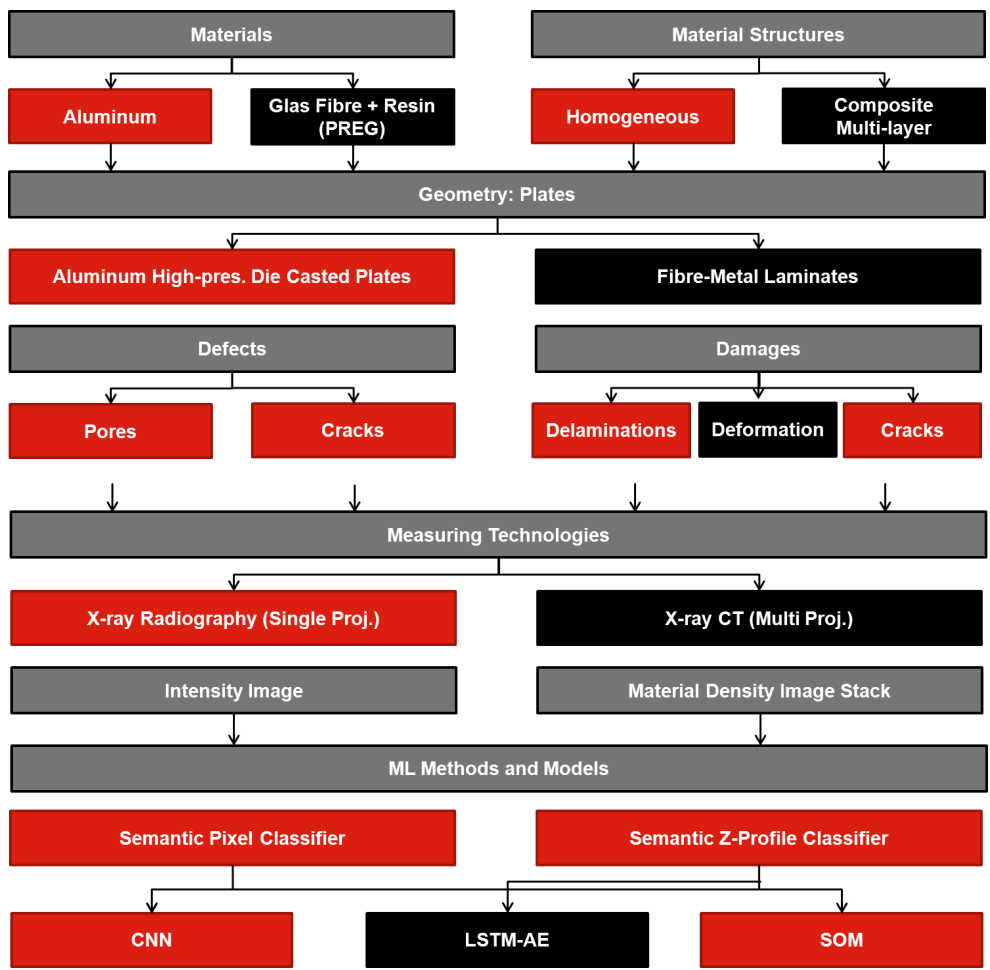

*Fig. 7. Taxonomy of damages, materials, devices, and feature detection methods applied to measuring data.*

## 6. Methods

The methods section distinguishes two data classes:

1. Single projection X-ray images (two-dimensional data);

2. Multi-projection X-ray images and CT material density reconstruction (three-dimensional data).

Furthermore, different method levels are addressed:

1. CAD Modeling of defects and damages in materials:

2. Simulation of synthetic X-ray images from CAD models (single- and multi-projections);

3. Filtered Back Projection for CT volume reconstruction;

4. Visualization;

5. Automated feature marking of defects or damages in 2D and 3D X-ray CT image stacks using different supervised and unsupervised data-driven methods, with a focus on semantic pixel classifier.

Finally, two different materials and defect classes are addressed:

1. Aluminum plates with pore defects;

2. Fibre-Metal Laminate plates with impact damages and secondary defects (delamination, cracks).

### 6.1 CAD Modelling

A CAD model is required to compute synthetic X-ray images. The CAD model consists of two parts:

1. The base-line structure and host material;

2. Several defects.

The CAD model is computed by using Constructive Solid Geometry (CSG) modeling. CSG provides union, difference, and intersection operations to construct arbitrarily structured and shaped materials.

A typical CSG program that is the input for the triangular mesh surface computation by using OpenSCAD [8] is shown in Ex. 1.

---

*Ex. 1. CSG example template: Homogeneous plate with elliptical pore defects.*

```
1: rotate ([90,90,90])
2: difference () {
3:   rotate ([90,0,0]) cube([100,4,40],true);
4:   union () {
```

```
 5:   translate([9.24,2.06,-0.97])
 6:    rotate ([0,0,-1.88])
 7:     scale([0.96,0.86,1.31])
 8:      sphere(r=0.5,$fn=20);
 9:    translate([3.15,-11.02,-0.55])
10:     rotate ([0,0,-7.34])
11:      scale([1.18,1.67,0.68])
12:       sphere(r=0.5,$fn=20);
13:    ...
14:    translate([-5.65,6.79,0.93])
15:     rotate ([0,0,64.48])
16:      scale([0.86,2.11,0.49])
17:       sphere(r=0.5,$fn=20);
18:  }
19:  }
```

Even complex materials, e.g., Fibre Metal Laminates, can be modeled accurately, e.g., creating single fibres. An example of a synthetic FML with 10000 glass fibres (embedded in resin) is shown in Fig. 8. The micro-scale structure of the fibre matrix is visible in the X-ray images with Moiré patterns. The computational time for the triangular mesh-grid transformation performed by OpenSCAD is about 10 minutes with 10000 fires and depends on the triangular approximation settings (number of segments to approximate circles and ellipses).

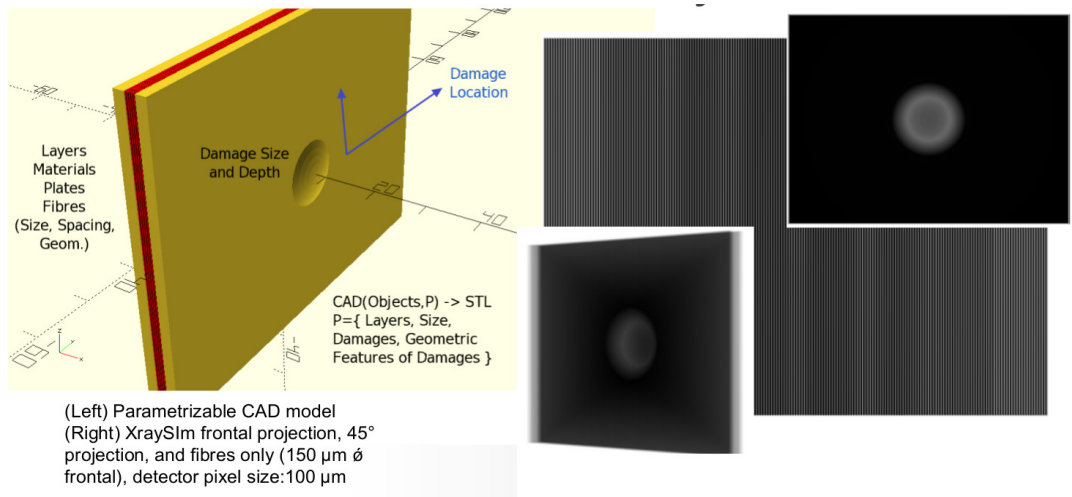

*Fig. 8. Complex FML model with single fibre modeling and resulting computed X-ray images (here with a simplified impact damage).*

Besides the low data variance rom real measure images and specimens, the second major challenge in data-driven modelling is the missing ground truth required to a create a strong training data set. The CAD model-based approach together with the synthetic data generation provides accurate annotated training examples.

### 6.2  X-ray Simulation

The X-ray simulation is used to create synthetic X-ray image projections from CAD models. The input is a polygon mesh grid (STL, Stereo lithography file format) model. A  STL file describes a raw, unstructured triangulated surface. A decomposition of multi-material structures in single density parts (finally merged in the simulator) is required for composite materials.

The 3D Model design deploys Constructive Solid Geometry (CSG) to create arbitrary complex structures (including Fibre structures). The output is a two-dimensional matrix providing the X-ray intensity image with a specific detector resolution (number of pixels) and pixel size, either floating point or integer data format (at least 16 Bits) are supported. A typical simulated X-ray image for a homogeneous aluminum plate with pore defects is shown in Fig. 9.

Spatial source, object, and detector geometries can be fully parametrized including rotated planes. The core software library used is *gvxr / gVirtualXray* using GPU computations and the OpenGL Shading Language (faster than 1ms / image) [2]. It is a simple ray tracer based on the  Beer-Lambert absorption law to compute the absorption of light (i.e. photons) by 3D objects (here polygon meshes). No reflection or scattering is considered, but the accuracy compared with real images is still high, demonstrated for medical applications in [6]. It is a C++ simulation library integrated in our simulator program *XraySim* [7].

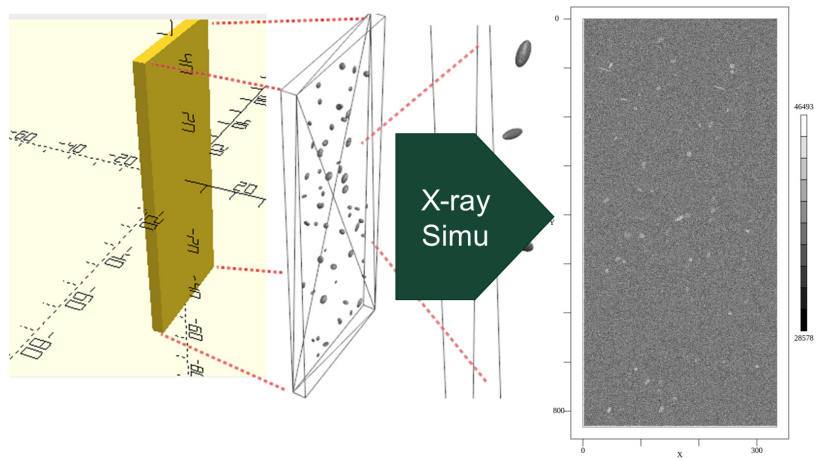

*Fig. 9. X-ray image projection simulation using a triangular multi-material mesh grid derived from a CAD model*

### 6.3  Pore Analysis in Die Casted Aluminum and Radiography Data

It is a challenge to estimate pore shapes (geometry, size), density, spatial distribution, and to distinguish reconstructed pores from image artifacts and noise.

Commonly a manual characterization by measuring the shape parameters of selected pores (e.g., using ImageJ [9] or IMEAS [10] analysis software) with ellipse approximation is performed. The goal is an automated pore analysis is desired split into three steps:

1. Semantic feature marking in the measured input images;

2. Application of point clustering methods to group marked pixels in groups;

3. Elliptical (2D) or ellipsoid (3D) approximation by iterative fitting algorithms.

Pores can be identified in reconstructed tomography 3D volumes from multi-projection X-ray measurements providing computed material density distributions with additionally geometric image processing reducing noise, although, artifacts, small pore sizes, and low density contrast can make this task difficult. Fig. 10 compares the accumulated semi-transparent top view of a reconstructed 3D volume (providing material density distributions) with original single-projection X-ray images from radiography, showing clearly the challenge to identify hidden pores in single X-ray images. The 3D volume view visualization intensifies regions of correlated homogeneous density, but damps noisy pixels (and very small defects with respect to the image resolution) in small uncorrelated regions by using pixel density clustering algorithms.

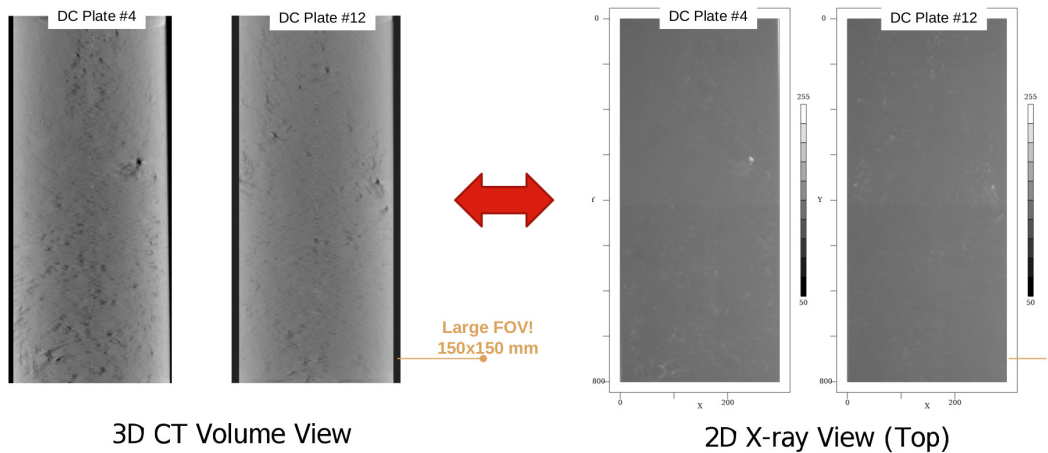

*Fig. 10. Comparison of layer-accumulated 3D CT material density views of plate specimens containing hidden pores and single absorption intensity X-ray images (seen from the top view of plate)*

The filtered back projection and CT reconstruction from rotational multi-projection measurements creates geometrical distortions with respect to contrast and density interpretation especially in boundary regions, e.g., blurring or smearing effects, making accurate geometric characterization of defects in comparison with radiography images difficult.

Therefore, it is a challenge (a) deriving geometric properties of the defects from 3D CT data (additionally requiring high measuring times) with respect to original X-ray single-projection images, and (b) direct feature marking of pores in single-projection X-ray images.

### 6.4 Pore Defect Feature Marking using a Semantic Pixel Classifier

A pixel classifier is a simple classification model applied to small segments of an image. The classifier model maps each pixel of an input image to a discrete classification value in an output feature map image.

The input is an X-ray image, the output is a feature map image of same size that marks pores, which is the base for geometric parameter and position estimations of pores, as shown in Fig. 11.

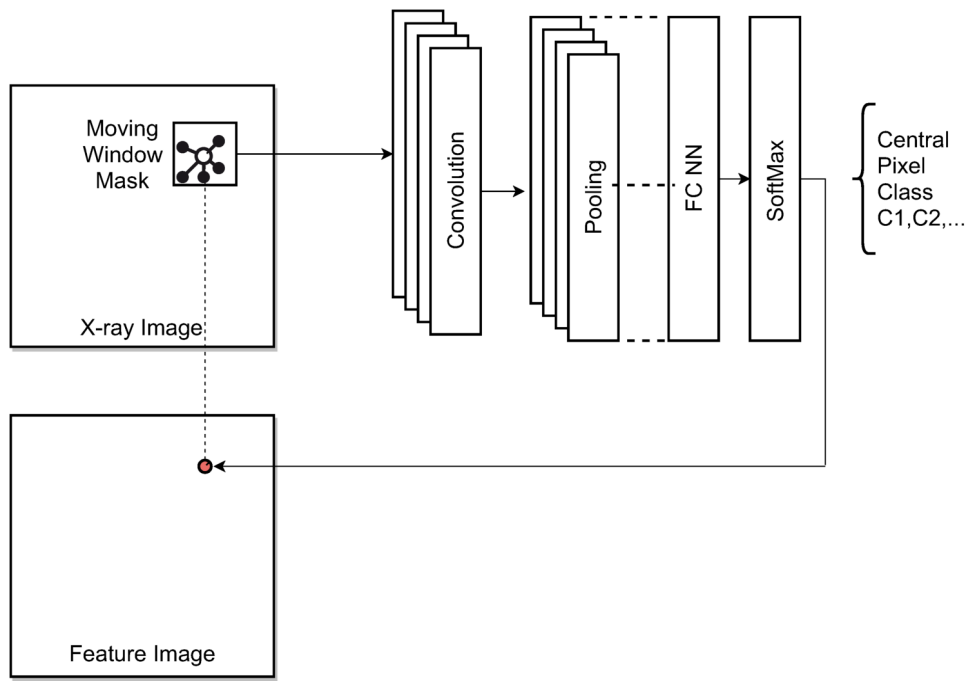

*Fig. 11. Semantic pixel classifier applied to X-ray single projection images to detect and mark hidden defects*

A pixel classifier is commonly implemented with a Convolutional Neural Network (CNN), mostly with only one or two convolution-pooling layer pairs. The input of the CNN is a sub-window masked out from the input image at a specific center position $(x, y)$. The output is a class (or a real value in the range [0,1] as an indicator level for a specific class). The neighboring pixels determine the classification result. The window and its CNN application is moved over the entire input image producing the respective feature output image (divide-and-conquer approach). Although, the single pixel classification is fast (computational time below 100 µs), the computation of the entire feature image can require a few seconds. Since the computations for each feature pixel is independent from each other, the task can be massively parallelized, finally using a cellular automaton processing architecture.

The pixel classifier is trained supervised with labelled regions of pores (i.e., pixels inside a closed polygon path surrounding a pore in the image), performing negative training (feature is the known difference from a base-line), as discussed in Sec. 4.

The semantic pixel classifier only provides a marking of image pixels as an intermediate latent feature. Point clustering (e.g., using the DBSCAN algorithm) can be used to group pixels marked with the same class to a extract list of geometric objects, e.g., pores, damages, and so on.

### 6.5 Workflow for Pore Characterization

The entire work flow for the pore defect analysis is shown in Fig. 12. It is a database driven system architecture. 3D CT data calculated from multi-projection measurements using a classical filtered back-projection algorithm is used to extract a core base set of pore parameters, passed to the stochastic Monte Carlo simulation process creating different CAD models of plates with pore defects. The CAD models are transformed in triangular mesh grids, finally used to compute synthetic X-ray images.

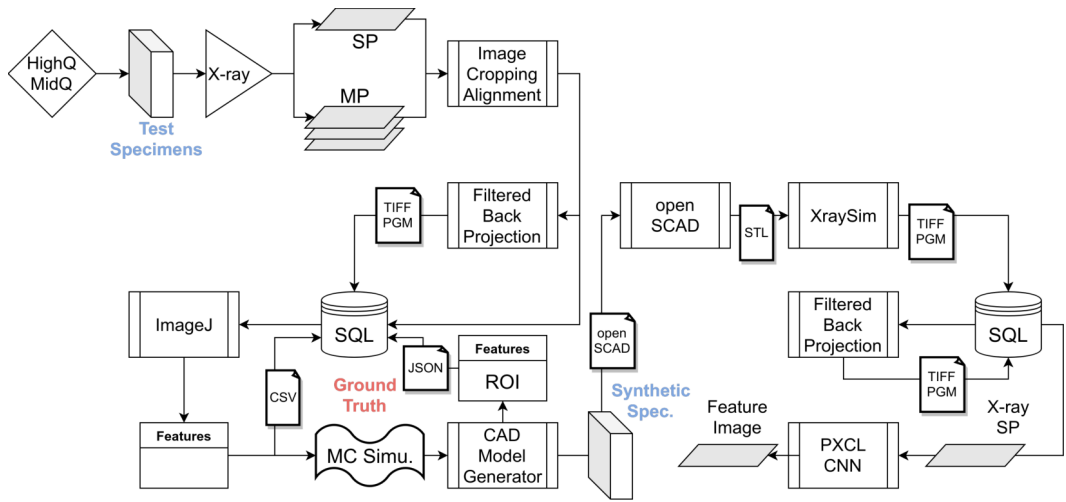

*Fig. 12. Details of the entire work flow used for the task of pore characterization in aluminum die casting plates.*

The synthetic data is used to train the pixel classifier, finally applied to measure single projection X-ray radiography images. The training is supervised and a negative training method, i.e., defects and defect classes are known in advance, and the predictive model is trained on the defects (negative features).

### 6.6 Damage Feature Marking in CT Data

The goal of the feature marking of damages in FML plates using X-ray images is two-folded:

1. Automated feature marking in high-resolution 3D CT data;

2. Automated feature marking in single projection radiography X-ray images (medium resolution).

The feature marling primarily aims to annotate images and image volumes with ROI boxes, finally enabling automated damage and defect characterization. In contrast to the pore defect marking as a known defect problem, the feature marking in the 3D CT data is originally an anomaly detection problem. An anomaly detector is typically a positive training problem, i.e., the model is trained on the ground truth base line (no defects) without knowing defects and damage characteristics in advance. Details can be found in [5].

The goal is to find (mark) damages (deformation, cracks, delaminations) in 3D CT volumes recorded from composite materials. The used method is called Z-Slicing of 3D CT volumes, as shown in Fig. 13. The Z-slices (one-dimensional depth signals) are finnaly passed to an anomaly detector to detect an in-depth anomaly in the x-y (top surface) plane.

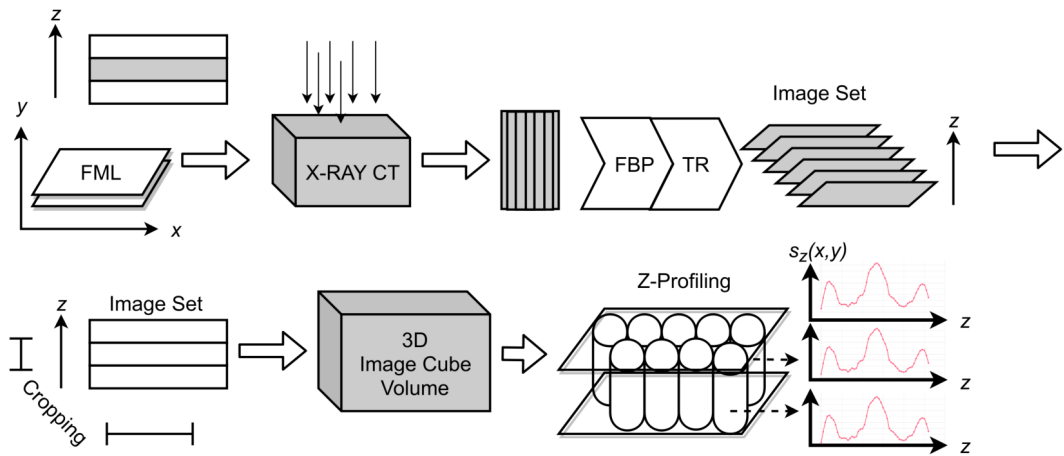

*Fig. 13. X-ray CT acquisition using FBP algorithms from ulti-projection X-ray images and applying Z-slicing (profiling) to the reconstrcuted 3-dim CT volume*

The z-slices are typically averaged by square x-y average filters, i.e., the i-th z-value in a z-slice is averaged over a squared region of pixels in the i-th image slice of the CT volume, providing a spatial correlation in the x-y plane, too. It is assumed that relevant defects are changing the z-profile structure, e.g., delaminations will stretch the z-structure locally (change in z-profile frequencies).

An anomaly detector can be build with a generative Autoencoder, either using a CNN or a LSTM-ANN, as shown in Fig. 14. A CNN handles a z-slice as a one-dimensional images, whereas a LSTM-ANN handles the z-slice as an ordered data series. The AE is trained with z-profile slices without defects or damages (base-line, ground truth data). The AE „learns" the z-profile structure of the FML plates and outputs a simplified representation (positive Training). If there is a damage/defect, the AE is not able to reconstruct the base-line structure, and an error occurs with respect to the input z-slice signal. The Mean Squared Error is computed and applied to a threshold detector. The output is an anomaly marker.

Alternatively, a CNN can be trained with damaged z-profiles to classify damaged versa undamged z-profile slices (classificator), i.e., performing negative training, as shown in Fig. 15.

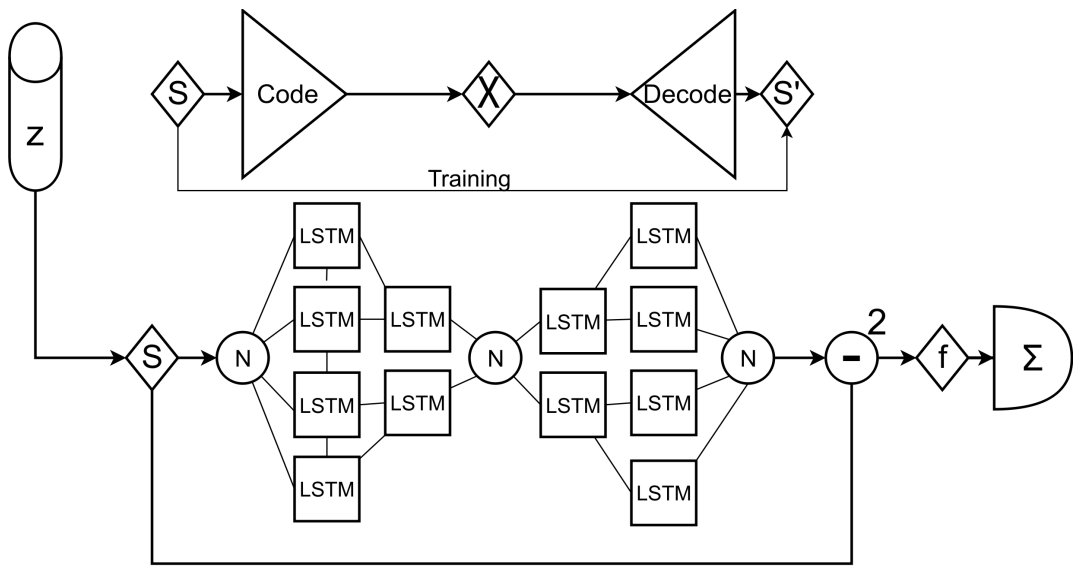

*Fig. 14. LSTM-ANN-based anomaly detector applied to z-slice profiles (positive training)*

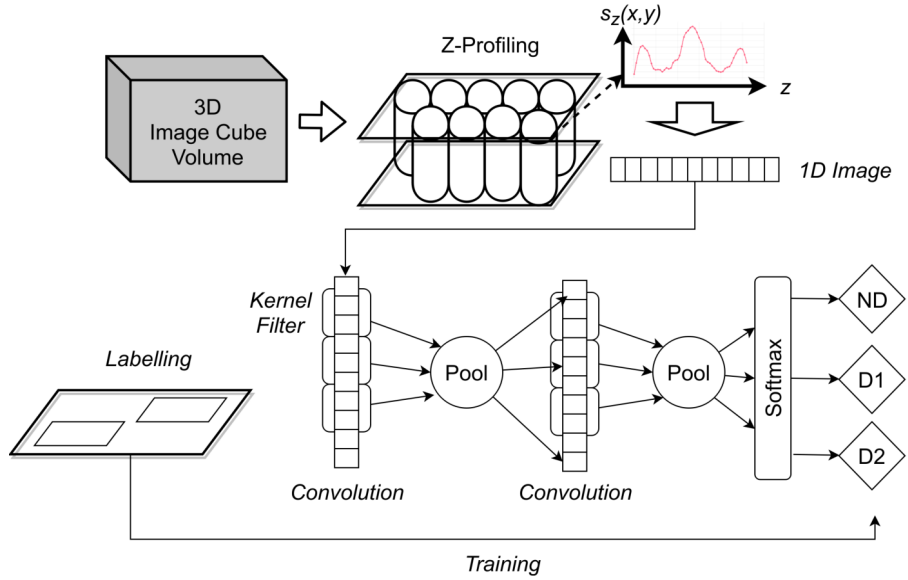

*Fig. 15. CNN-based damager detector applied to z-slice profiles (negative training)*

This use-case poses the same data constraints as the pores use-case, but under tougher conditions. In this work, multi-layer composite materials (Fibre-Metal laminates, FML) with impact damages were used. Each specimen provides one impact damage. A typical sample set contains less than 10 different specimens, each with a distinct and unique impact damage (and base-line = no damage). Therefore, X-ray image simulation and CT reconstruction should be used to generate an extended synthetic data set. But the CAD modeling of damages and the FML itself is a challenge and still work under progress.

Manual marking must be performed to model the compiste material and damages by drawing boundary polygon lines (between material layers, i.e., metal-fibre,fibre-metal, metal-air), as shown in Fig. 16. Edge detection by using a Canny edge filter can aid to mark boundaries and damage outlines. Damages in FML plates are cracks (in fibre layer) and delaminations. The markings must be added to a set of z-slices to get 3D and volume information. Canny edge detection can be used in conjunction with an additional manually triggered pixel to polygon clustering to semi-automatically create ROI polygon paths.

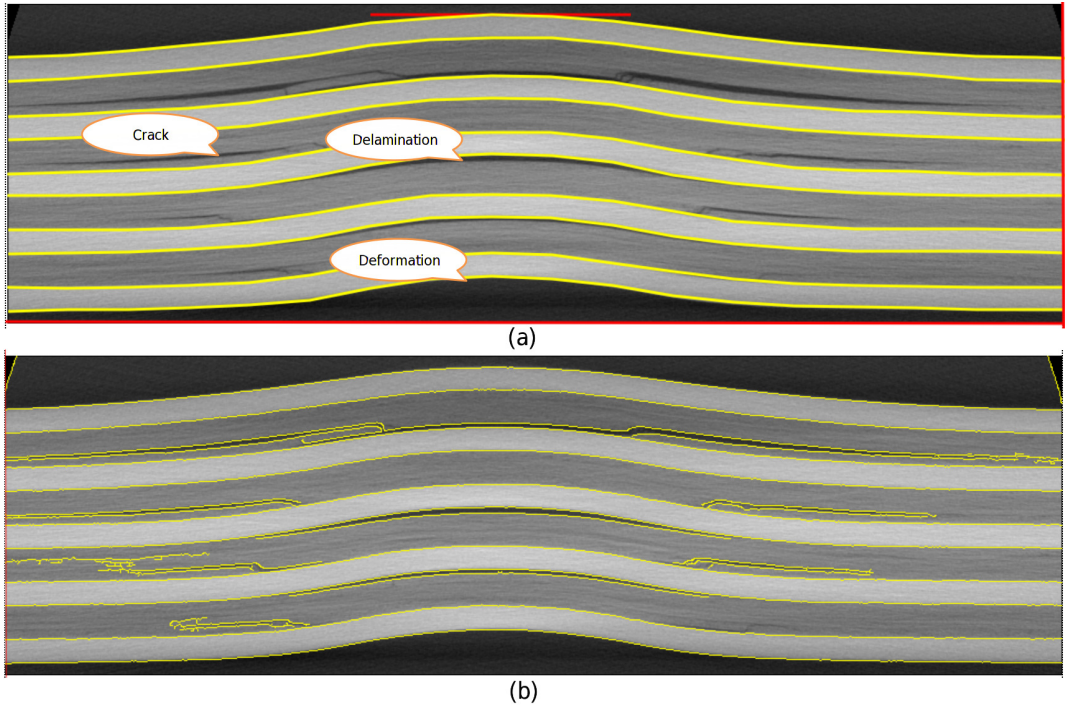

*Fig. 16. (a) Manually marked layer boundaries using polygon lines (b) Canny edge filter overlay*

## 7. Results

### 7.1 Pore Analysis in Die Casted Aluminum and Radiography Data

The first experiment using measured single-projection X-ray radiography data (MidQ device) was performed with training data retrieved from manual ROI annotation with visual inspection. The previously described semantic pixel classifier was trained with about 1000 training segments. The manual ROI marking of pores is a challenged due to the low contrast and noise. Therefore,, the labeling error rate is estimated with at least 30% (FP and FN). But the pixel classifier was able to find most of the marked pores. Because there is no ground truth, no statistical analysis of test data can be performed (at least with a significant statistical strength).

The second experiment uses synthetic training and test data with ground truth labeling. The synthetic X-ray images were computed from CAD models created with a Monte Carlo simulation. The base parameter data for typical pore characteristics and spatial distributions were extracted from 3D CT data from real specimens (using the the MidQ device and our own filtered back-projection algorithm with less post-filtering intervention). Two synthetic plates with about 100 pores were created. The computed X-ray images were superimposed with Gaussian (detector) noise with a relative σ=10%.

The pixel classifier, a classical CNN with an architecture shown in Alg. 1, was trained with about 1000 training example segments from the synthetic data set only using a common SGD trainer. Less than 100 epochs were required for stable prediction. A Due to the increased noise and noise sensitivity the training set was imbalanced with about 30% pore and 70% background segments. The TP rate (detecting a pixel of a pore) was about 95%, the FN rate was about 5% using synthetic X-ray images computed form the two synthetic plates, as shown in Fig. 17. Besides the good feature marking results, there is a strange observation. Two independent plates and images with different pore distributions, different gaussian pixel noise, produce same artifacts (upper right corner of both images)! There is actually no rigorous explanation, but it is assumed that numerical noise from the X-ray simulator is the cause. This observation shows a specific noise sensitivity of the feature detector not known in advance ( Fig. 17 c).

```
{type='input',size=[sgementSize,segmentSize,1]},
{type='conv',kernel.size=[5,5],
 filter=filterA,stride=1,padding=2,activation='relu'},
{type='pool',kernel.size=[2,2],stride=2},
{type='conv',kernel.size=[5,5],
 filter=filterB,stride=1,padding=2,activation='relu'},
{type='pool',kernel.size=[2,2],stride=2},
{type='softmax',num.classes=2}
```

*Alg. 1. Parameterizable CNN pixel classifier with parameter setting [sgementSize-filterA-filterB]*

The trained model was then applied to real measured single-projection X-ray images using the Mid-Q and Low-Q devices. Some selected results are shown in Figures 18 and 19 for the Mid-Q and Low-Q devices, respectively, but using the same feature marking model. Due to the missing ground truth of the real specimens no statistical analysis can be performed, just an estimation by visual expert inspection that the results are promising and that the detected pores, the spatial distribution, and their orientation correlated with the manufacturing process and known observations.

The feature marking in the Low-Q images (Low-quality device, but with higher spatial resolution compared with the Mid-Q device) shows a higher marking rate and an increased marking of smaller pores. This result raises doubts and a differential diagnosis was performed using a pore-free rolled aluminum plate, as shown in Fig. 20. The feature detector still marks non-existing pores, depending on the average intensity level (exposure time). Repeated experiments show different pore markings. Compared with the die casting specimen detection rate we assume an error of about 30% (FP). The Low-Q devices poses a significant higher non-Gaussian noise, probably with some spatial correlation. The feature detectors seems to be sensitive to this kind of noise (although, the training was performed with Gaussian noise). Again, the FN error rate is unknown.

Another observation from Fig. 20 is remarkable. The marked pores resulting just from intensity noise in the X-ray images show an aligned orientation like in the results from the die casting plates. We can conclude that there is a training data bias (pore orientation and alignment) inherited by the feature marking model, i.e., the sensitivity is anisotropically biased.

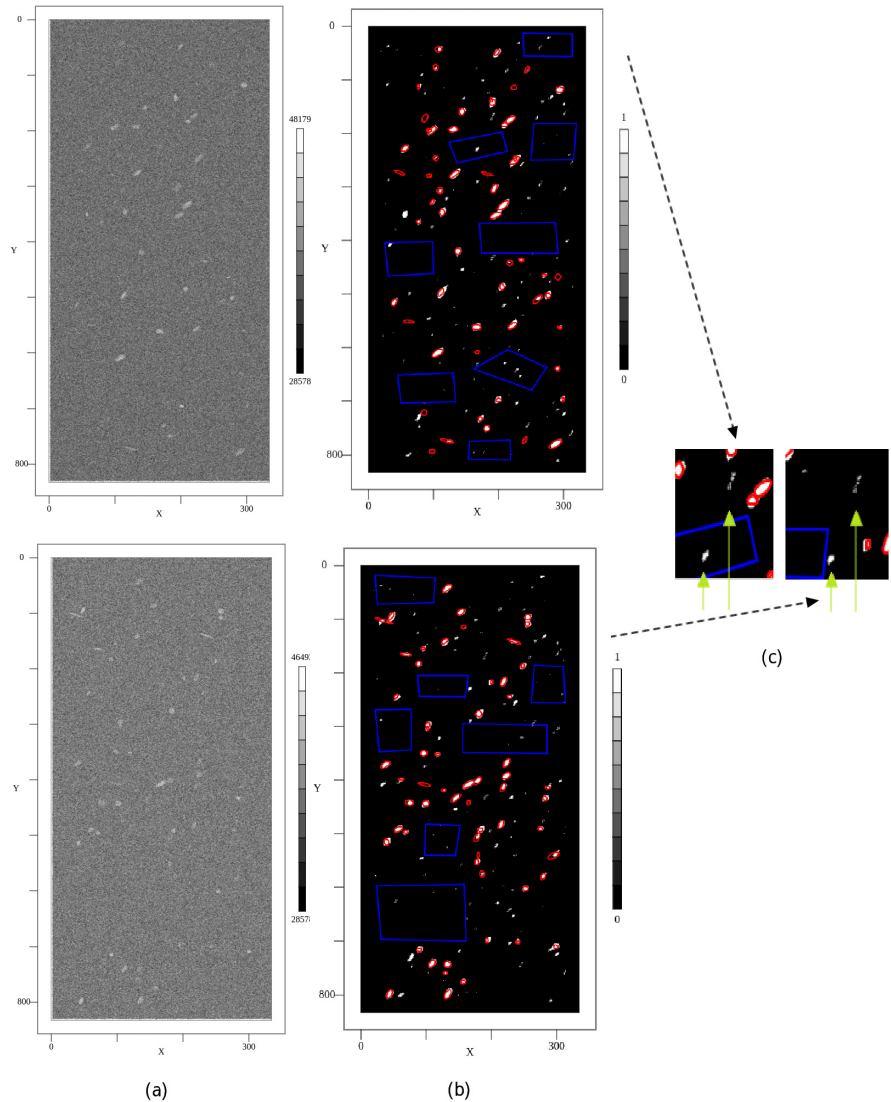

Fig. 17. (a) Single-projection X-ray radiography images computed by XraySim (M=2, pixel size 150µm 1000x1000 pixels, cropped) using the **synthetic plates** (b) Feature marking results of CNN pixel classifier predicted from the images using a [20-8-8] CNN with an overlay of red pore ROIs from CAD model, extended with blue background areas (c) Same artifacts found in both images

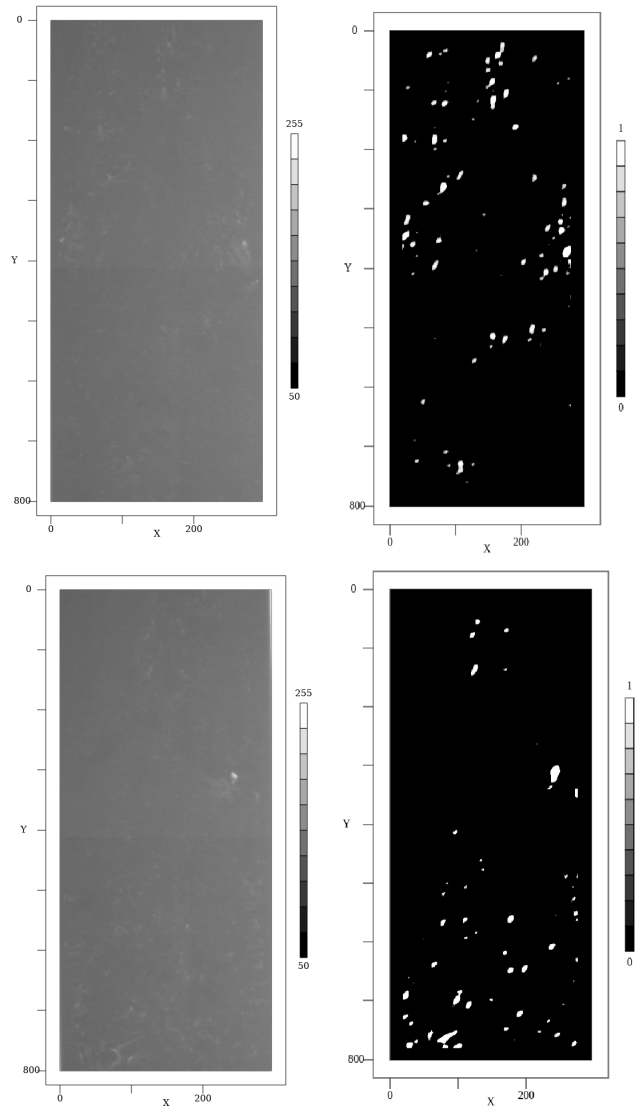

*Fig. 18. (Left) Single projection X-ray radiography images measured with the Mid-Q device with a 150 × 150 mm FOV, two different plates (Right) Feature marking results of CNN pixel classifier predicted from the images using a [20-8-8] CNN*

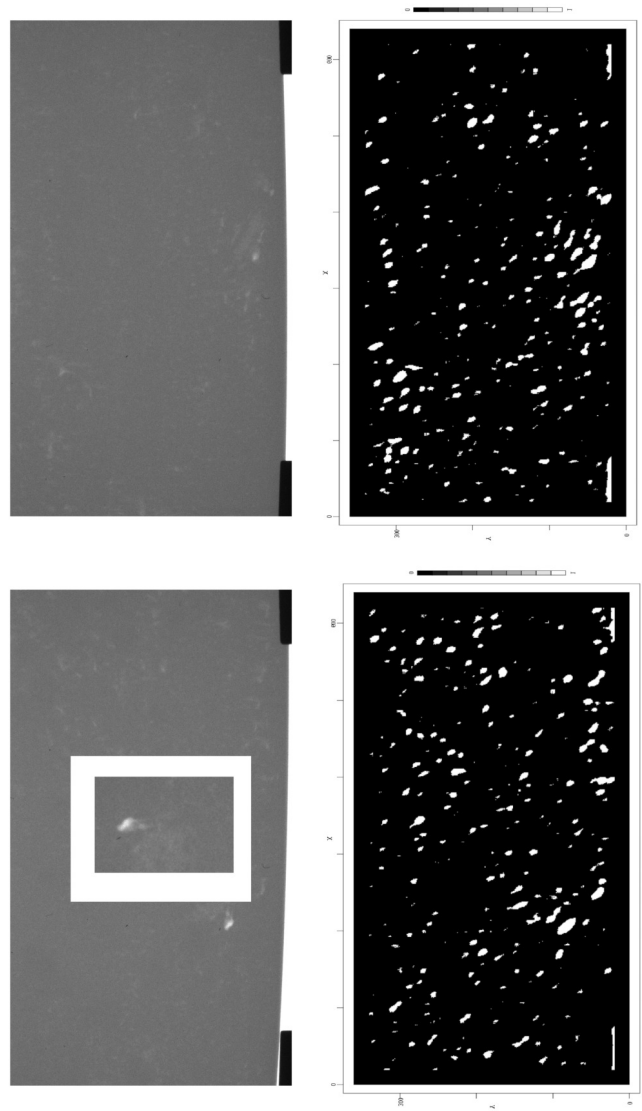

*Fig. 19. (Left) Single projection X-ray radiography images measured with the Low-Q device with a 80 × 40 mm FOV, two different plates (Right) Feature marking results of CNN pixel classifier predicted from the images using a [20-8-8] CNN*

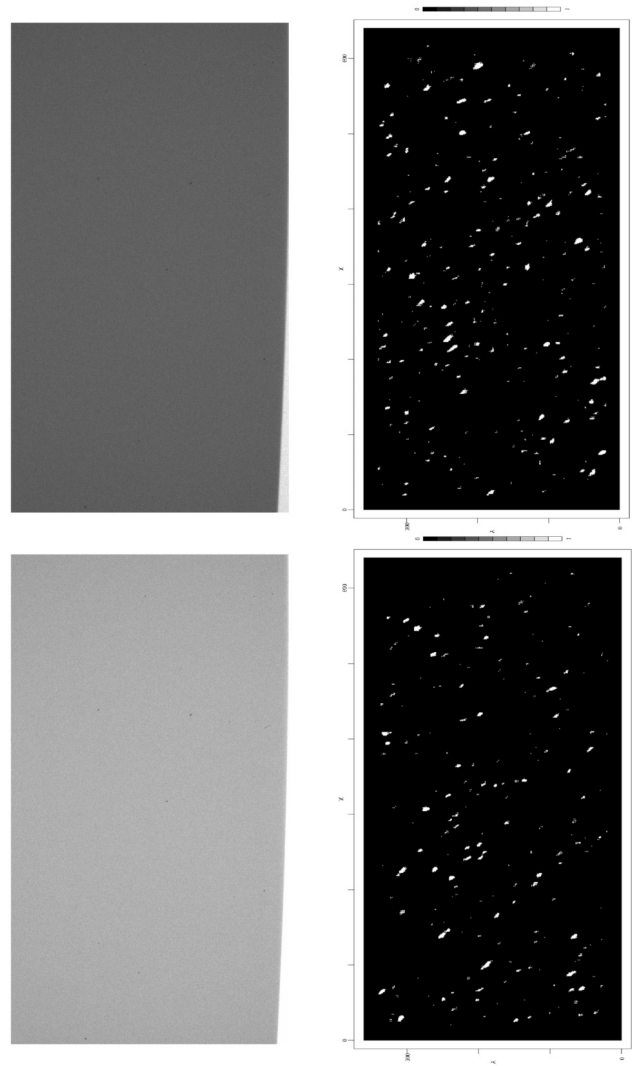

*Fig. 20. (Left) Single projection X-ray radiography images measured with the Low-Q device with a 80 × 40 mm FOV of a pore-free rolled aluminum plate with different exposure time (Right) Feature marking results of CNN pixel classifier predicted from the image using a [8-8-4] CNN showing noise markings (FP)*

### 7.2 Damage Feature Marking in FML and CT Data

The investigation of feature marking in non-homogeneous materials like Fibre-Metal laminates is still preliminary. There aim is to detect typical hidden damages from impact deformations by using single-projection X-ray images, too.

To create synthetic training data, a CAD model must be created. The CAD model consists of two domains:

1. The undamaged host material, here a sandwich structure of alternating aluminum or steel plates and fibre-resin (PREG) layers;

2. The damages (impact deformation, delamination of layers, cracks).

The geometric characteristics of the layers itself can be either assumed by design or from polylines extracted by visual manual inspection from 3D CT image slices measured with the Micro-CT High-Q device (resolution and accuracy about 50 microm). Geometric distortions resulting from the CT reconstruction and from the measuring device itself are currently ignored.

The modeling of the defects is a challenge. The impact deformation and inter-layer delaminations can only be modeled by poly-line extrusion. Cracks and inner-layer defects must be modeled by subtractive geometry. Neither delamination nor hidden cracks increase the overall material volume (mass). But subtractive geometry must be density (material) preserving, which is it not. Therefore, there must be a density and volume proof of a damaged plate compared with an undamaged plate, still a challenge.

Using the *imeas* tool [10] providing edge detctors (Canny) and a semi-automatic pixel tracking it was possible to derive:

1. Layer boundary polylines;

2. Damage polygon paths (delamination and cracks).

Examples are shown in Fig. 21. The ROI parameters can be saved in a CSV table, finally processed by the plate and damage CSG CAD model generator.

The synthetic CAD model generation of the deformed host material (due to impact damages) is rather simple. It can be achieved with extrusion operations using the layer profile polylines. The modeling of the secondary damages (cracks, delamination) is much more complicated because both damage classes overlap spatially. And as described above the material volume and mass sum must be kept constant (compared with undamaged specimen material). This can be solved by an iterative process monitoring the entire material volume from the created triangular surface mesh grids.

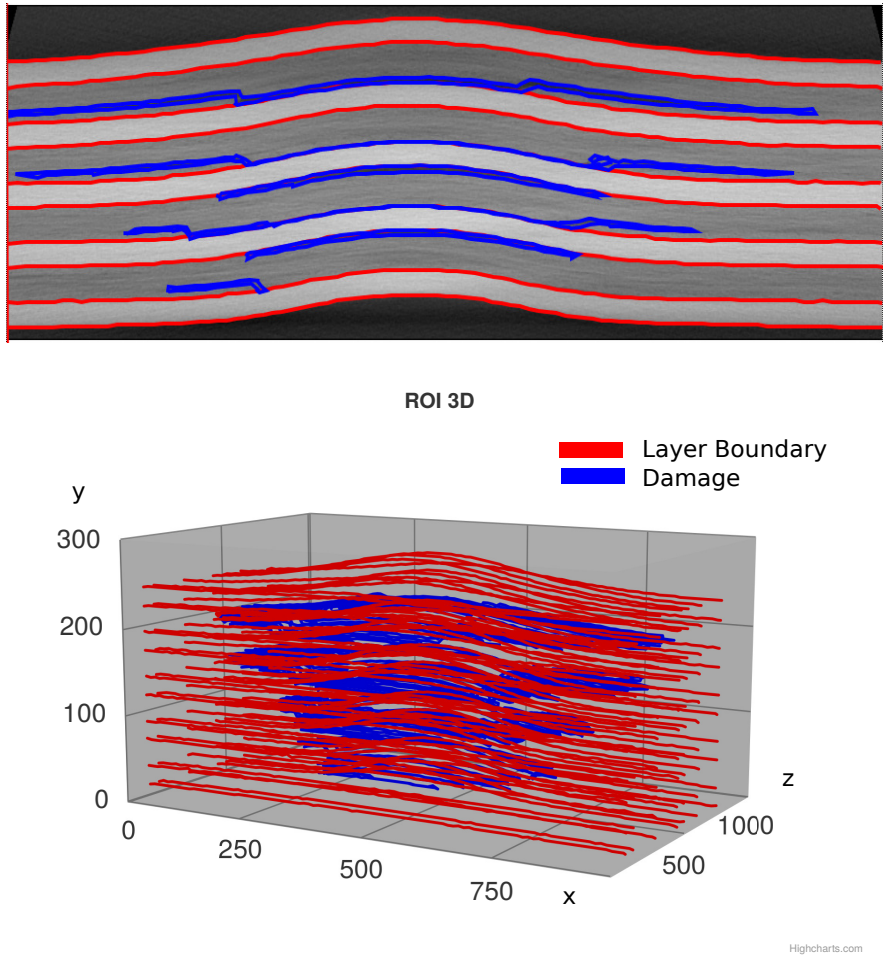

*Fig. 21. Example of ROI marking in 3D CT image slice volumes from an impact damaged FML plate with material density images from μCT High-Q device (Top) One slice with ROI polygon markings (Bottom) 3D ROI view*

The training of damage and anomaly detectors applied to 3D CT image volumes using synthetic data is still an ongoing work.

Results retrieved from AE and CNN models trained with measured data taken from [5] are shown inf Figures 22 and 23. The simple z-slice predictor models are able to detect impact damage deformations and pseudo defects (e.g., embedded foils or delaminations by resin wash-out).

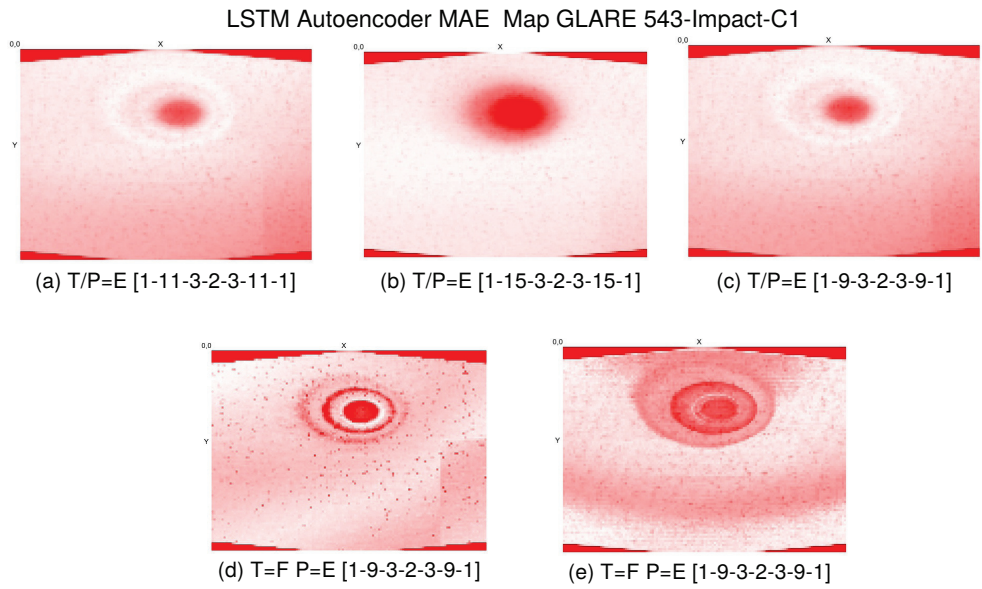

(a) T/P=E [1-11-3-2-3-11-1]  (b) T/P=E [1-15-3-2-3-15-1]  (c) T/P=E [1-9-3-2-3-9-1]

(d) T=F P=E [1-9-3-2-3-9-1]  (e) T=F P=E [1-9-3-2-3-9-1]

*Fig. 22. AE-MAE feature map of reconstructed AE X-ray CT data of specimen with impact damage (a-c): Training and Prediction using selected areas from impacted specimen, (e-f): Training with undamaged specimen, prediction with impacted specimen; in brackets: AE network layer structure*

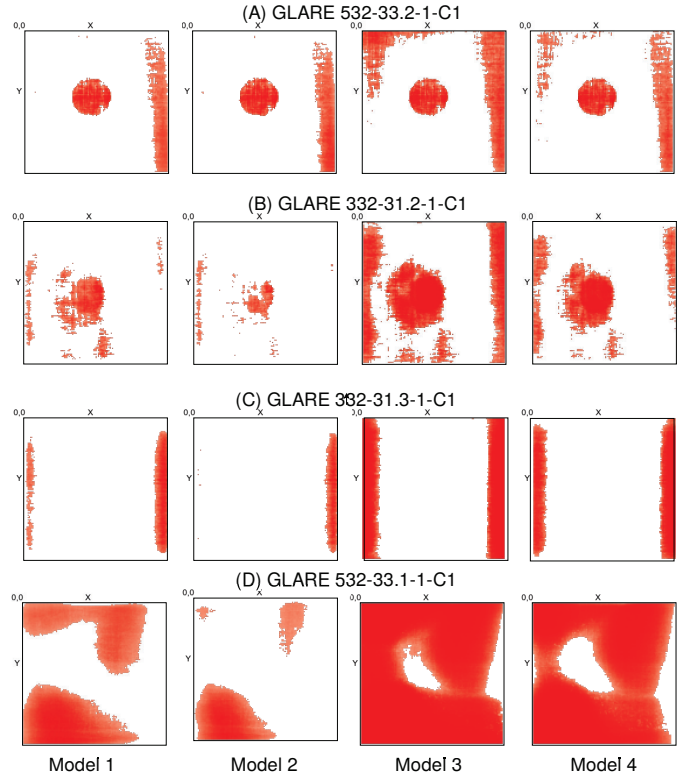

*Fig. 23. Damage feature maps retrieved from four different CNN classifiers and for the specimen A (training and prediction), B, C, and D. All specimens contained pseudo defects like resin wash-out (A,B) and delaminations (C,D)*

## 8. Conclusions

This work addresses damage diagnostics in hybrid and composite materials like Fibre-metal laminates (FML) as well as homogeneous materials like aluminum die casting plates using X-ray diagnostics. Both single- and multi-projection data (Radiography and CT) are considered. Data-driven ML models with mid and low complexity are applied either to raw X-ray intensity image or reconstructed material density volume data. CNN as well as AE-LSTM based networks could be used to identify defects or anomalies. The models implement Semantic pixel classifiers. The comparison of results from Mid- and High-Q devices with a Low-Q and low.cost devices showed that the degraded quality has impact of the prediction and feature marking results, but that they are still usable.

Due to the missing data and experimental variance synthetic X-ray images were computed from generated CAD models. Defects like pores can be easily modeled by CSG, primary and secondary damages resulting from impact damages are still hard to be modeled correctly.

The goal is to train the predictor models with synthetic data only providing a ground truth labeling not available in real measured images. In the case of pore detection using single-projection X-ray intensity images this approach succeeded.

## 9. References

[1]  S. Khan, H. Rahmani, S. A. A. Shah, and M. Bennamoun, A Guide to Convolutional Neural Networks for Computer Vision. Morgan & Claypool Publishers, 2018

[2]  gvirtualxray, https://gvirtualxray.fpvidal.net, accessed on-line on 24.1.2023

[3]  F. P. Vidal, Introduction to X-ray simulation on GPU using gVirtualXRay, In Workshop on Image-Based Simulation for Industry 2021 (IBSim-4i 2020), London, UK, October, 2021

[4]  X-Ray Projection Simulator based on Raytracing, https://git.scc.kit.edu/dach/raytracingx-rayprojectionsimulator, accessed on-line on 24.1.2023

[5]  C. Shah, S. Bosse, and A. von Hehl, "Taxonomy of Damage Patterns in Composite Materials, Measuring Signals, and Methods for Automated Damage Diagnostics," Materials, vol. 15, no. 13, p. 4645, Jul. 2022, doi: 10.3390/ma15134645.

[6]  Jamie Lea Pointon, Tianci Wen, Jenna Tugwell-Allsup, Aaron Sújar, Jean Michel Létang, and Franck Patrick Vidal, Simulation of X-ray projections on GPU: Benchmarking gVirtualXray with clinically realistic phantoms, Computer Methods and Programs in Biomedicine 234 (2023)

[7]  XraySim: The X-ray Simulator, https://github.com/bslab/xraysim, accessed on-line 1.10.2023

[8]  OpenSCAD, https://openscad.org/, on-line, accessed 1.10.2023

[9]  ImageJ, https://imagej.net/ij/, on-line accessed on-line 1.11.2023

[10]  IMEAS, Stefan Bosse, The Image Measurement Too, https://github.com/bsLab/imeas, accessed on-line 1.11.2023

[11]  S. Bosse, Automated Damage and Defect Detection with Low-Cost X-Ray Radiography using Data-driven Predictor Models and Data Augmentation by X-Ray Simulation, 10th International Electronic Conference on Sensors and Applications (ECSA), MDPI, 15-30.11.2023